# A NOVEL APPROACH FOR HANDWRITTEN DEVNAGARI CHARACTER RECOGNITION

Sandhya Arora[1]
Latesh Malik [2]
Debotosh Bhattacharjee[2]
Mita Nasipuri[3]

[1]Meghnad Saha Institute of Technology, Kolkata
[2]G.H. Raisoni College of Engineering, Nagpur
[2]University Of Calcutta, Kolkata
[3]Jadavpur University, Kolkata
sandhyabhagat@yahoo.com lateshmalik@rediffmail.com
debotoshb@hotmail.com mitanasipuri@yahoo.com

**ABSTRACT**

*In this paper a method for recognition of handwritten devanagari characters is described. Here, feature vector is constituted by accumulated directional gradient changes in different segments, number of intersections points for the character, type of spine present and type of shirorekha present in the character. One Multi-layer Perceptron with conjugate-gradient training is used to classify these feature vectors. This method is applied to a database with 1000 sample characters and the recognition rate obtained is 88.12%*

## 1. Introduction

In this paper our concern is devanagari script. It is the script for Hindi which is official language of India. It is also the script of Sanskrit, Marathi, and Nepali languages. More than 450 million people on the globe use the script. Sinha et al.[8] and some other researchers [9] have reported various aspects of devanagari script recognition. However none of the works have considered handwritten devanagari characters. Devanagari has 11 vowels and 33 simple consonants. Besides the consonants and the vowels, other constituent symbols on devanagari are set of vowel modifiers called matra which can be placed on the left, right, top or at the bottom of a character or conjunct, pure-consonants (half letters) which when combined with other consonants yield conjuncts. This method can be described in three steps: preprocessing (discussed in section 2), feature extraction (section 3), classification (section 4) and the results are given in section 5.

## 2. Preprocessing

The preprocessing steps remove any noise, distortions in the input character and convert the character in a form recognizable by the system. It consists of the following steps :-

**2.1 Size Scaling:-** Approximate dimension of the character is determined and a tight fit rectangular boundary is formed around the character. All characters are scaled to 140 X 140 matrix.

**2.2 Thinning, pruning and noise removal: -** Thinning of binary pattern consists of successive deletion of dark points (i.e. changing them to white points) along the edges of the pattern until it is thinned to a line. Let ZO(P1) count be the number of zeros to nonzero transitions in the ordered set P2,P3,P4,P5,P6,P7,P8,P9,P2. Let Nzcount(P1) be the number of non zero neighbours of P1.

| | | |
|---|---|---|
| P3 | P2 | P9 |
| P4 | P1 | P8 |
| P5 | P6 | P7 |

Then P1 is deleted if
Step 1: 2<= Nzcount <=6
Step 2: and ZO(P1)=1
Step 3:and P2.P4.P8=0 or ZO(P2) != 1
Step 4: and P2.P4.P6=0 or ZO(P4) != 1

The procedure[7],all above said steps are repeated until no further changes occur in the image. The corner positions are the special cases to be considered in the thinning procedure. This results some redundant pixels. To remove this redundancy we have to apply certain masks given below. And result of applying these masks in Figure 2

| | | | |
|---|---|---|---|
| - 1 -<br>1 1 1<br>- 1 - | Don't remove center pixel | - 1 -<br>1 1 -<br>- - 0 | Remove center pixel |
| - 1 -<br>- 1 1<br>0 - - | Remove center pixel | - - 0<br>1 1 -<br>- 1 - | Remove center pixel |
| 0 - -<br>- 1 1<br>- 1 - | Remove center pixel | | |

**Table 1 Masks**

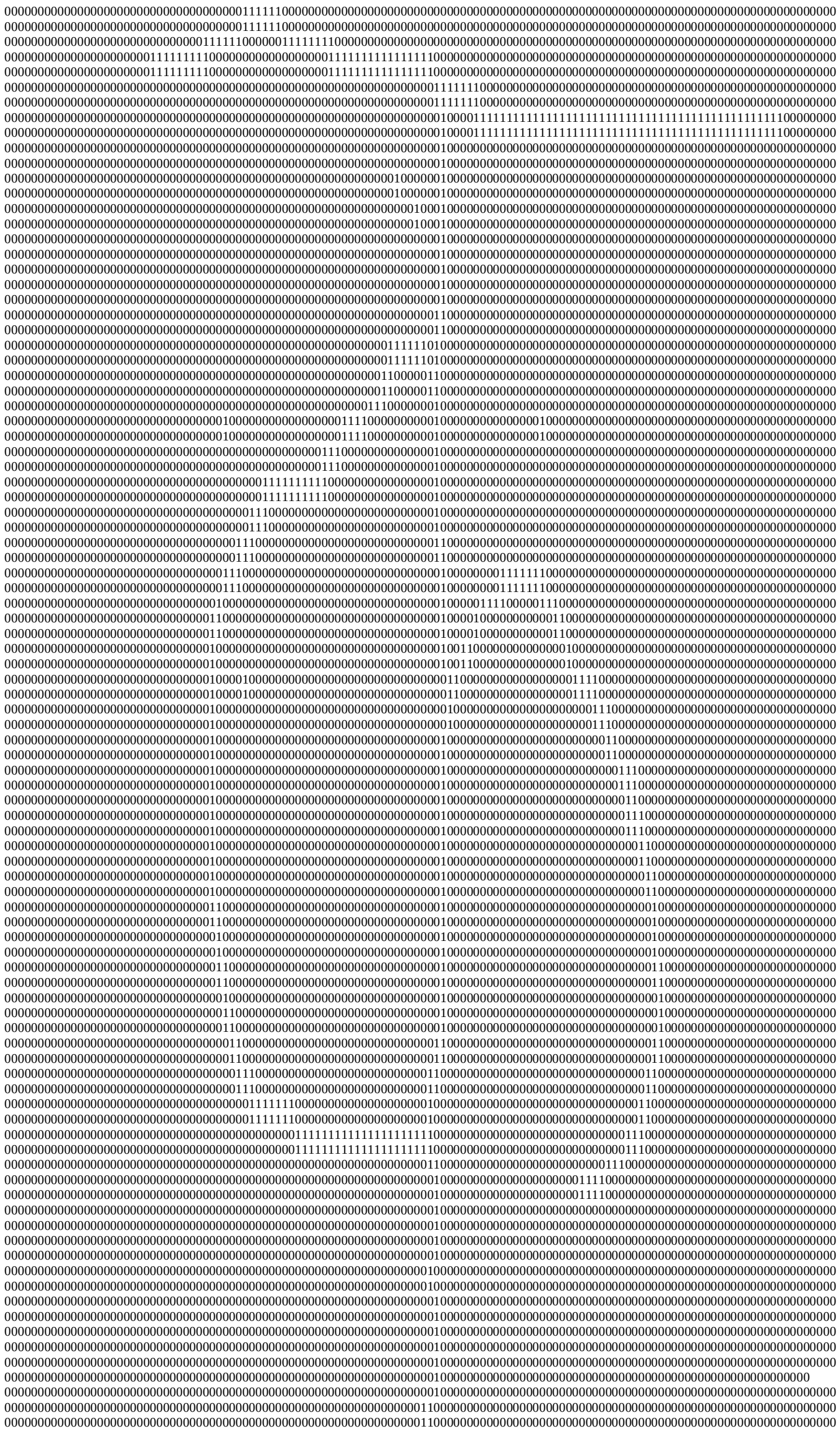

**Figure1 Result with redundant pixels in image**

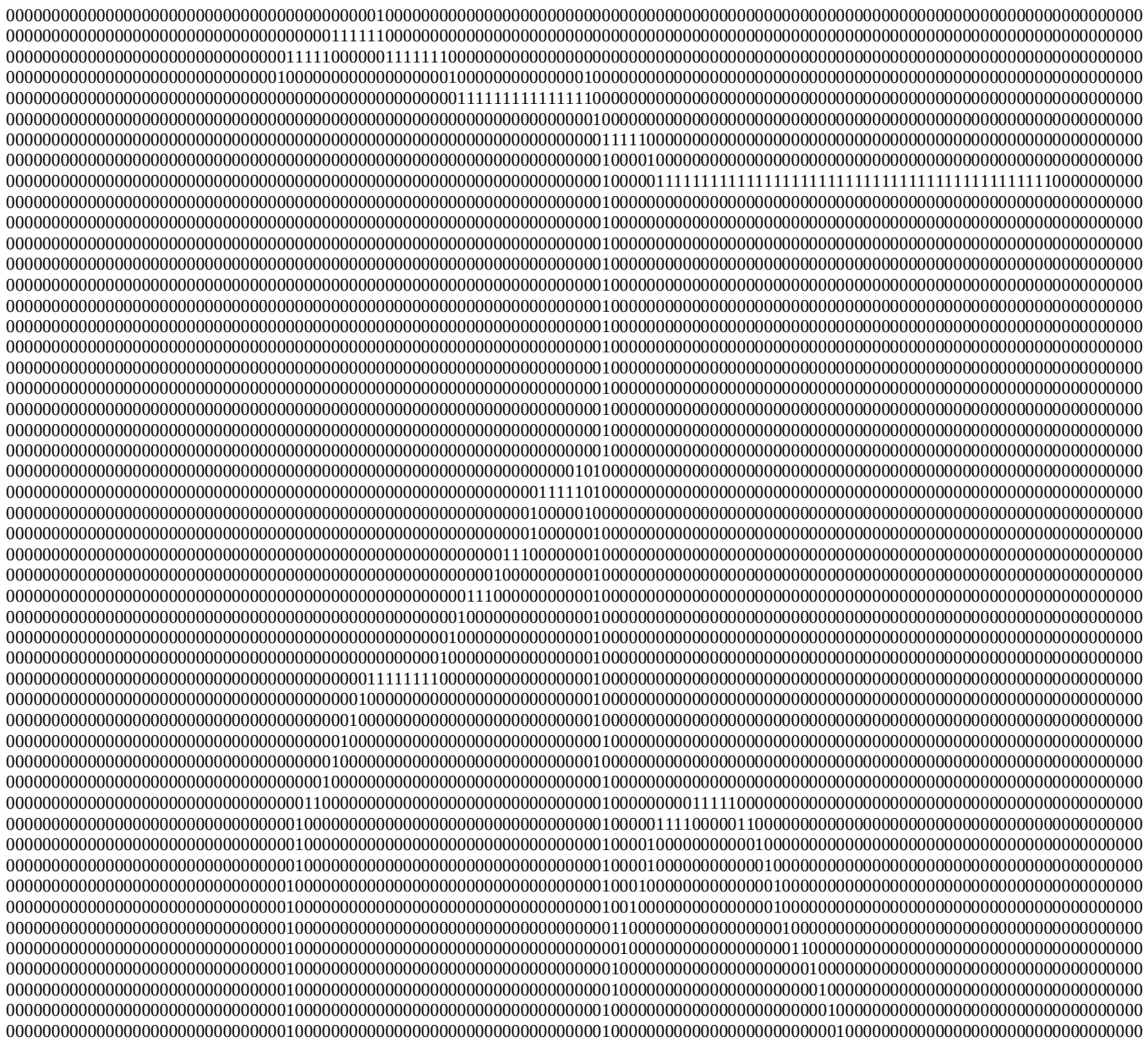

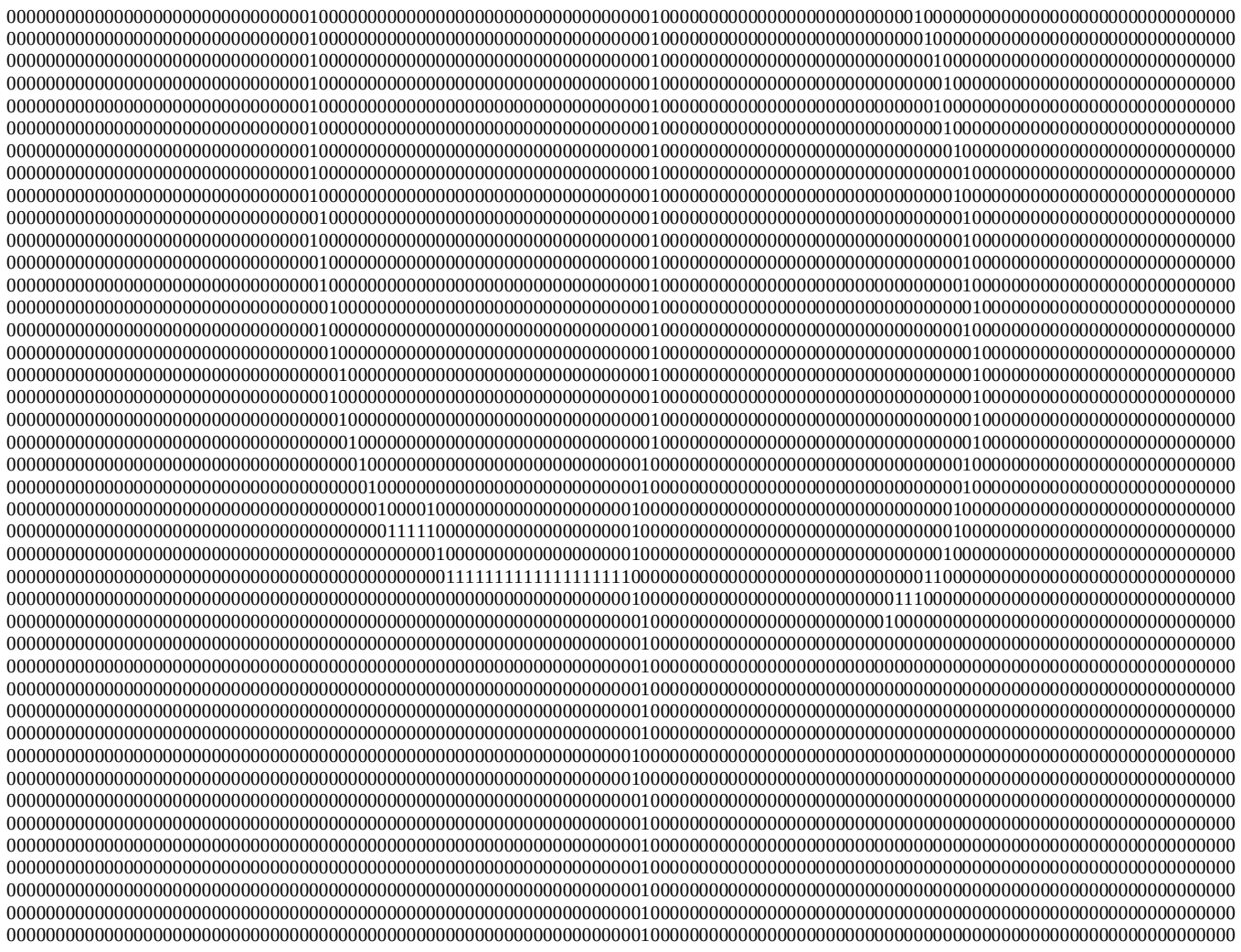

**Figure2 One-pixel-wide skeleton without noise after applying mask**

## 3 Feature extraction

Following features are extracted in this step:-

**3.1 Shirorekha detection:-** It is assumed that the first pixel available for a given character when looked from right to left and is above 1/3 part of the character is the shirorekha i.e nothing can go past the shirorekha as is the normal practice. The last pixel is traced out for the character and the shirorekha is detected by priority based neighbourhood search.

**3.2 Vertical Spine detection:-** The vertical spine detected as the downward near straight line looking from right of each character. A straight line to qualify as a spine should be at least ¾ of the height of the character. Type of spine (present in mid or at end of the character or no spine present ) is also detected.

**3.3 Number of intersections:-** Intersection points are calculated using shirorekha and vertical spine and other pixel present the character.

**3.4 Accumulated directional Gradient Change feature detection:-** The source image is skeletonized and divided into different segments(4, 9,16,25 segments). For each segment calculate the accumulated gradient change using following steps[1,4]:-

Step 1: Consider two successive rows at a time. The whole process should be performed in horizontal manner.

Step 2: Let $B_i$ be the first black pixel found in row i

Step 3: Calculate the distance between $B_i$ and $B_i$+1

Step 4: Sum up the distances(gc's) for whole segment

These gc's values may be zero, negative, positive (Fig 3). It may contribute zero for horizontal or vertical segements, may contribute positive for convex shapes, negative for concave shapes. The idea is that, in practice although for two different images, a few of the segments may yield identical gc values, it would be rare if all the components in the vector were same for completely different shapes. The complete image's gradient feature can be described with a component vector V=(gc1,gc2…..gc4) or V=(gc1,gc2…..gc9) or V=(gc1,gc2….gc16) or V=(gc1,gc2….gc25.

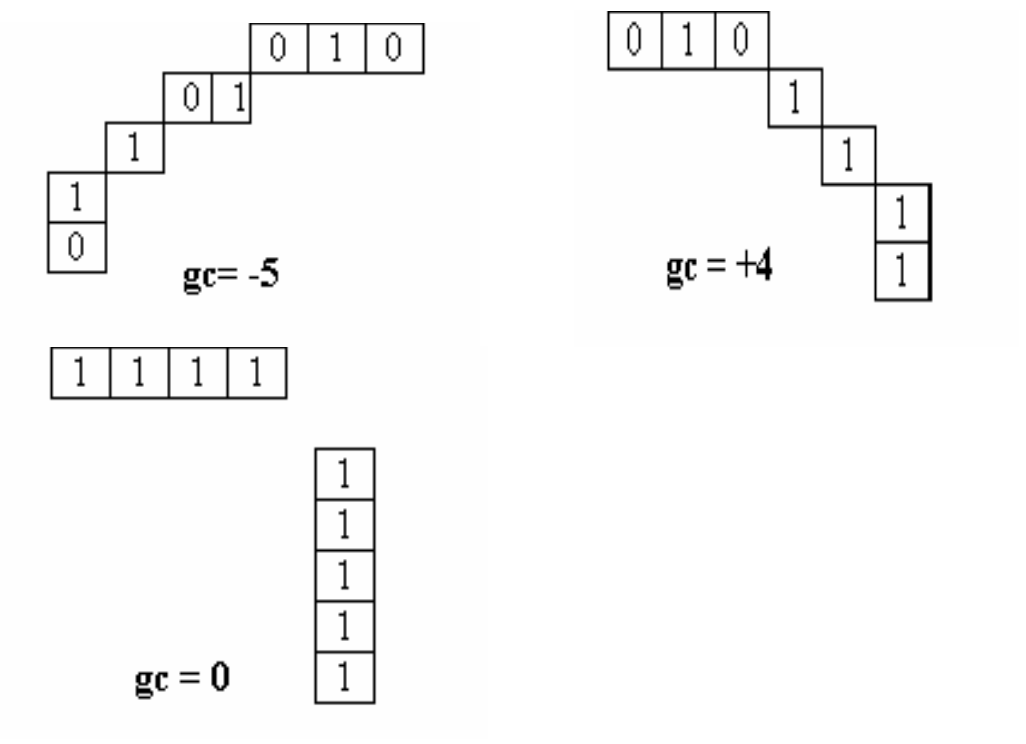

**Figure 3 gc Computation**

Using all above features , feature vectors are computed for the whole character image.

## 4. Classification

### 4.1 Grouping based on Feature:-

4.1.1 **Shirorekha type:-** Some characters contain a shirorekha throughout, while others contain a partial or no shirorekha. Thus character can be in one of the three category.

**4.1.2 Vertical Spine Location :-** Character can be divided into three category according to this aspect – Spine present at end , spine at mid or No spine present.

For classification of feature vectors of character that belongs to a particular category one Multi-layer Perceptron(MLP) has been designed. Apart from input layer and output layer it consists of one hidden Layer. Input to the Neural network are the accumulated gradient feature, vertical spine type, presence/absence of shirorekha type, number of intersections in image. Number of neurons depends on the segmentation of the image for calculating gradient feature. The network was trained with the normalized data using conjugate-gradient( CG) method of training. This method was preferred to the gradient descent method, since CG takes into account the non-linearity of the surface. The CG procedure does not ask user to specify any parameters such as learning rate.

The conjugate gradient algorithm start out by searching in the steepest descent direction (negative of the gradient) on the first iteration.

$$\mathbf{p_0 = -g_0} \quad \text{------------------} \ (1)$$

A line search is then performed to determine the optimal distance to move along the current search direction:

$$\mathbf{x_{k+1}} = \mathbf{x_k} + \boldsymbol{\alpha}_\mathbf{k}\, \mathbf{p_k} \quad \text{---------------------------} \ (2)$$

Then the next search direction is determined so that it is conjugate to previous search directions. The general procedure for determining the new search direction is to combine the new steepest descent direction with the previous search direction:

$$\mathbf{p_k = -\, g_k + \boldsymbol{\beta}_k p_{k-1}} \quad \text{------------------------------} \ (3)$$

where $\mathbf{x_k}$ is a vector of current weights and biases, $\mathbf{g_k}$ is the current gradient, and $\boldsymbol{\alpha}_\mathbf{k}$ is the learning rate.

## 5. Result and Conclusion

Different classifiers have been used for handwritten digit recognition, such as statistical[3], structural and neural networks[5], 99.77% recognition rate have been reported for handwritten digits using multiple features & multiple neural networks[2] but for characters 93% recognition rate have been reported for printed devanagari characters.

We have simulated the present recognition scheme on 25 characters database of devanagari characters. This database includes 750 sample set for training set, 250 sample set for testing. A few samples are shown in Table 3. Ideal samples are shown in Table 2.

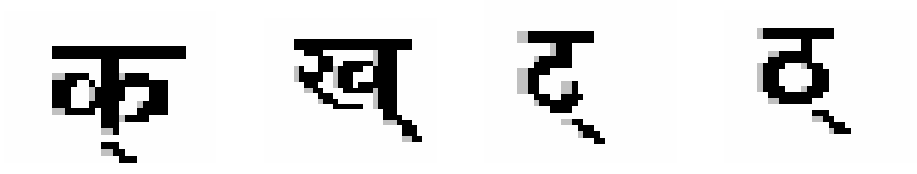

**Table 2 Ideal samples of devanagari characters**

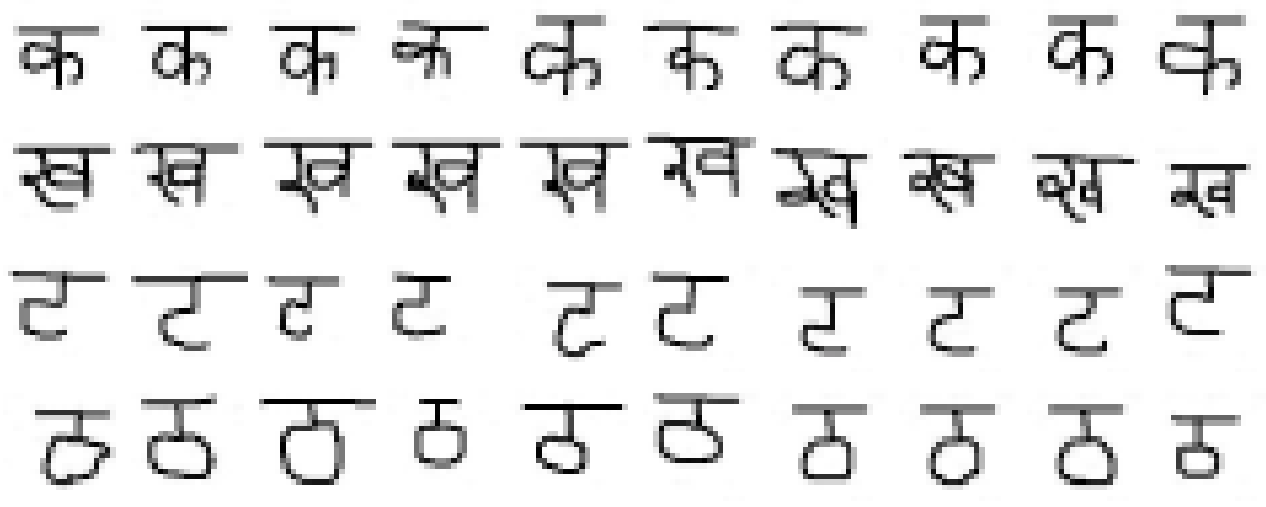

**Table 3 A typical sample data subsets of handwritten devanagari characters**

We divided character image into different size segments. For each segment we computed gradient changes. These gradient values are normalized for training and testing of Neural Network. For different size segment (input vector to Neural Network), we have made several simulation runs varying the normalization factor and we have observed that recognition accuracy on test set of samples can be improved by taking optimal segment size and it can also be slightly improved using proper normalization factor. We achieved 88.2% of accuracy for hand written devanagari characters.